# Sum-Product Networks: A New Deep Architecture

**Hoifung Poon** and **Pedro Domingos**
Computer Science & Engineering
University of Washington
Seattle, WA 98195, USA
{hoifung,pedrod}@cs.washington.edu

## Abstract

The key limiting factor in graphical model inference and learning is the complexity of the partition function. We thus ask the question: what are general conditions under which the partition function is tractable? The answer leads to a new kind of deep architecture, which we call *sum-product networks (SPNs)*. SPNs are directed acyclic graphs with variables as leaves, sums and products as internal nodes, and weighted edges. We show that if an SPN is *complete* and *consistent* it represents the partition function and all marginals of some graphical model, and give semantics to its nodes. Essentially all tractable graphical models can be cast as SPNs, but SPNs are also strictly more general. We then propose learning algorithms for SPNs, based on backpropagation and EM. Experiments show that inference and learning with SPNs can be both faster and more accurate than with standard deep networks. For example, SPNs perform image completion better than state-of-the-art deep networks for this task. SPNs also have intriguing potential connections to the architecture of the cortex.

## 1 INTRODUCTION

The goal of probabilistic modeling is to represent probability distributions compactly, compute their marginals and modes efficiently, and learn them accurately. Graphical models [22] represent distributions compactly as normalized products of factors: $P(X = x) = \frac{1}{Z} \prod_k \phi_k(x_{\{k\}})$, where $x \in \mathcal{X}$ is a $d$-dimensional vector, each *potential* $\phi_k$ is a function of a subset $x_{\{k\}}$ of the variables (its scope), and $Z = \sum_{x \in \mathcal{X}} \prod_k \phi_k(x_{\{k\}})$ is the *partition function*. Graphical models have a number of important limitations. First, there are many distributions that admit a compact representation, but not in the form above. (For example, the uniform distribution over vectors with an even number of 1's.) Second, inference is still exponential in the worst case. Third, the sample size required for accurate learning is worst-case exponential in scope size. Fourth, because learning requires inference as a subroutine, it can take exponential time even with fixed scopes (unless the partition function is a known constant, which requires restricting the potentials to be conditional probabilities).

The compactness of graphical models can often be greatly increased by postulating the existence of *hidden variables* $y$: $P(X = x) = \frac{1}{Z} \sum_y \prod_k \phi_k((x, y)_{\{k\}})$. Deep architectures [2] can be viewed as graphical models with multiple layers of hidden variables, where each potential involves only variables in consecutive layers, or variables in the shallowest layer and $x$. Many distributions can only be represented compactly by deep networks. However, the combination of non-convex likelihood and intractable inference makes learning deep networks extremely challenging. Classes of graphical models where inference is tractable exist (e.g., mixture models [17], thin junction trees [5]), but are quite limited in the distributions they can represent compactly. This paper starts from the observation that *models with multiple layers of hidden variables allow for efficient inference in a much larger class of distributions*. Surprisingly, current deep architectures do not take advantage of this, and typically solve a harder inference problem than models with one or no hidden layers.

This can be seen as follows. The partition function $Z$ is intractable because it is the sum of an exponential number of terms. All marginals are sums of subsets of these terms; thus if $Z$ can be computed efficiently, so can they. But $Z$ itself is a function that can potentially be compactly represented using a deep architecture. $Z$ is computed using only two types of operations: sums and products. It can be computed efficiently if $\sum_{x \in \mathcal{X}} \prod_k \phi_k(x_{\{k\}})$ can be reorganized using the distributive law into a computation involving only a polynomial number of sums and products. Given a graphical model, the inference problem in a nutshell is to perform this reorganization. But we can instead learn from the outset a model that is already in efficiently computable form, viewing sums as implicit hidden variables. This leads naturally to the question: what is the broadest class of models that admit such an efficient form for $Z$?

We answer this question by providing conditions for tractability of $Z$, and showing that they are more general than previous tractable classes. We introduce *sum-product networks (SPNs)*, a representation that facilitates this treatment and also has semantic value in its own right. SPNs can be viewed as generalized directed acyclic graphs of mixture models, with sum nodes corresponding to mixtures over subsets of variables and product nodes corresponding to features or mixture components. SPNs lend themselves naturally to efficient learning by backpropagation or EM. Of course, many distributions cannot be represented by polynomial-sized SPNs, and whether these are sufficient for the real-world problems we need to solve is an empirical question. Our experiments show they are quite promising.

## 2 SUM-PRODUCT NETWORKS

For simplicity, we focus first on Boolean variables. The extension to multi-valued discrete variables and continuous variables is discussed later in this section. The negation of a Boolean variable $X_i$ is represented by $\bar{X}_i$. The indicator function [.] has value 1 when its argument is true, and 0 otherwise. Since it will be clear from context whether we are referring to a variable or its indicator, we abbreviate $[X_i]$ by $x_i$ and $[\bar{X}_i]$ by $\bar{x}_i$.

We build on the ideas of Darwiche [7], and in particular the notion of *network polynomial*. Let $\Phi(x) \geq 0$ be an unnormalized probability distribution. The network polynomial of $\Phi(x)$ is $\sum_x \Phi(x) \Pi(x)$, where $\Pi(x)$ is the product of the indicators that have value 1 in state $x$. For example, the network polynomial for a Bernoulli distribution over variable $X_i$ with parameter $p$ is $px_i + (1-p)\bar{x}_i$. The network polynomial for the Bayesian network $X_1 \rightarrow X_2$ is $P(x_1)P(x_2|x_1)x_1x_2 + P(x_1)P(\bar{x}_2|x_1)x_1\bar{x}_2 + P(\bar{x}_1)P(x_2|\bar{x}_1)\bar{x}_1x_2 + P(\bar{x}_1)P(\bar{x}_2|\bar{x}_1)\bar{x}_1\bar{x}_2$.

The network polynomial is a multilinear function of the indicator variables. The unnormalized probability of evidence (partial instantiation of $X$) $e$ is the value of the network polynomial when all indicators compatible with $e$ are set to 1 and the remainder are set to 0. For example, $\Phi(X_1 = 1, X_3 = 0)$ is the value of the network polynomial when $\bar{x}_1$ and $x_3$ are set to 0 and the remaining indicators are set to 1 throughout. The partition function is the value of the network polynomial when all indicators are set to 1. For any evidence $e$, the cost of computing $P(e) = \Phi(e)/Z$ is linear in the size of the network polynomial. Of course, the network polynomial has size exponential in the number of variables, but we may be able to represent and evaluate it in polynomial space and time using a *sum-product network*.

**Definition 1** *A sum-product network (SPN) over variables*

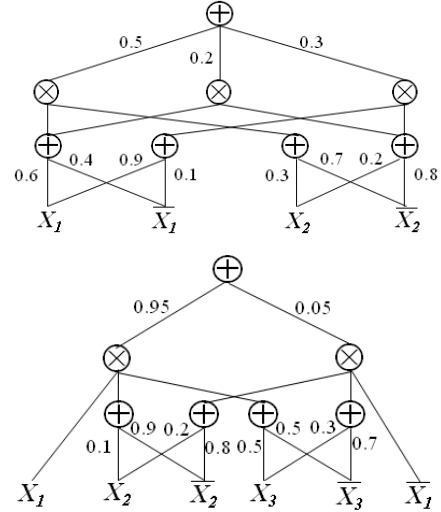

Figure 1: Top: SPN implementing a naive Bayes mixture model (three components, two variables). Bottom: SPN implementing a junction tree (clusters $(X_1, X_2)$ and $(X_1, X_3)$, separator $X_1$).

$x_1, \ldots, x_d$ *is a rooted directed acyclic graph whose leaves are the indicators $x_1, \ldots, x_d$ and $\bar{x}_1, \ldots, \bar{x}_d$ and whose internal nodes are sums and products. Each edge $(i, j)$ emanating from a sum node $i$ has a non-negative weight $w_{ij}$. The value of a product node is the product of the values of its children. The value of a sum node is $\sum_{j \in Ch(i)} w_{ij} v_j$, where $Ch(i)$ are the children of $i$ and $v_j$ is the value of node $j$. The value of an SPN is the value of its root.*

Figure 1 shows examples of SPNs. In this paper we will assume (without loss of generality) that sums and products are arranged in alternating layers, i.e., all children of a sum are products or leaves, and vice-versa.

We denote the sum-product network $S$ as a function of the indicator variables $x_1, \ldots, x_d$ and $\bar{x}_1, \ldots, \bar{x}_d$ by $S(x_1, \ldots, x_d, \bar{x}_1, \ldots, \bar{x}_d)$. When the indicators specify a complete state $x$ (i.e., for each variable $X_i$, either $x_i = 1$ and $\bar{x}_i = 0$ or $x_i = 0$ and $\bar{x}_i = 1$), we abbreviate this as $S(x)$. When the indicators specify evidence $e$ we abbreviate it as $S(e)$. When all indicators are set to 1, we abbreviate it as $S(*)$. The subnetwork rooted at an arbitrary node $n$ in the SPN is itself an SPN, which we denote by $S_n(.)$. The values of $S(x)$ for all $x \in \mathcal{X}$ define an unnormalized probability distribution over $\mathcal{X}$. The unnormalized probability of evidence $e$ under this distribution is $\Phi_S(e) = \sum_{x \in e} S(x)$, where the sum is over states consistent with $e$. The partition function of the distribution defined by $S(x)$ is $Z_S = \sum_{x \in \mathcal{X}} S(x)$. The scope of an SPN $S$ is the set of variables that appear in $S$. A variable $X_i$ appears negated in $S$ if $\bar{x}_i$ is a leaf of $S$ and non-negated if $x_i$ is a leaf of $S$.

For example, for the SPN in Figure 1, $S(x_1, x_2, \bar{x}_1, \bar{x}_2) = 0.5(0.6x_1 + 0.4\bar{x}_1)(0.3x_2 + 0.7\bar{x}_2) + 0.2(0.6x_1 + $

$0.4\bar{x}_1)(0.2x_2+0.8\bar{x}_2)+0.3(0.9x_1+0.1\bar{x}_1)(0.2x_2+0.8\bar{x}_2)$. The network polynomial is $(0.5 \times 0.6 \times 0.3 + 0.2 \times 0.6 \times 0.2 + 0.3 \times 0.9 \times 0.2)x_1x_2 + \ldots$ If a complete state $x$ is $X_1 = 1, X_2 = 0$, then $S(x) = S(1,0,0,1)$. If the evidence $e$ is $X_1 = 1$, then $S(e) = S(1,1,0,1)$. Finally, $S(*) = S(1,1,1,1)$.

**Definition 2** *A sum-product network $S$ is* valid *iff $S(e) = \Phi_S(e)$ for all evidence $e$.*

In other words, an SPN is valid if it always correctly computes the probability of evidence. In particular, if an SPN $S$ is valid then $S(*) = Z_S$. A valid SPN computes the probability of evidence in time linear in its size. We would like to learn only valid SPNs, but otherwise have as much flexibility as possible. We thus start by establishing general conditions for the validity of an SPN.

**Definition 3** *A sum-product network is* complete *iff all children of the same sum node have the same scope.*

**Definition 4** *A sum-product network is* consistent *iff no variable appears negated in one child of a product node and non-negated in another.*

**Theorem 1** *A sum-product network is valid if it is complete and consistent.*

*Proof.* Every SPN $S$ can be expressed as a polynomial $\sum_k s_k \prod_k(\ldots)$, where $\prod_k(\ldots)$ is a monomial over the indicator variables and $s_k \geq 0$ is its coefficient. We call this the *expansion* of the SPN; it is obtained by applying the distributive law bottom-up to all product nodes in the SPN, treating each $x_i$ leaf as $1x_i + 0\bar{x}_i$ and each $\bar{x}_i$ leaf as $0x_i + 1\bar{x}_i$. An SPN is valid if its expansion is its network polynomial, i.e., the monomials in the expansion and the states $x$ are in one-to-one correspondence: each monomial is non-zero in exactly one state (condition 1), and each state has exactly one monomial that is non-zero in it (condition 2). From condition 2, $S(x)$ is equal to the coefficient $s_x$ of the monomial that is non-zero in it, and therefore $\Phi_S(e) = \sum_{x \in e} S(x) = \sum_{x \in e} s_x = \sum_k s_k n_k(e)$, where $n_k(e)$ is the number of states $x$ consistent with $e$ for which $\Pi_k(x) = 1$. From condition 1, $n_k = 1$ if the state $x$ for which $\Pi_k(x) = 1$ is consistent with the evidence and $n_k = 0$ otherwise, and therefore $\Phi_S(e) = \sum_{k : \Pi_k(e)=1} s_k = S(e)$ and the SPN is valid.

We prove by induction from the leaves to the root that, if an SPN is complete and consistent, then its expansion is its network polynomial. This is trivially true for a leaf. We consider only internal nodes with two children; the extension to the general case is immediate. Let $n^0$ be an arbitrary internal node with children $n^1$ and $n^2$. We denote the scope of $n^0$ by $V^0$, a state of $V^0$ by $x^0$, the expansion of the subgraph rooted at $n^0$ by $S^0$, and the unnormalized probability of $x_0$ under $S_0$ by $\Phi^0(x^0)$; and similarly for $n^1$ and $n^2$. By the induction hypothesis, $S^1 = \sum_{x^1} \Phi^1(x^1)\Pi(x^1)$ and $S^2 = \sum_{x^2} \Phi^2(x^2)\Pi(x^2)$.

If $n^0$ is a sum node, then $S^0 = w_{01} \sum_{x^1} \Phi^1(x^1)\Pi(x^1) + w_{02} \sum_{x^2} \Phi^2(x^2)\Pi(x^2)$. If $V^1 \neq V^2$, then each state of $V^1$ (or $V^2$) corresponds to multiple states of $V^0 = V^1 \cup V^2$, and therefore each monomial from $V^1$ ($V^2$) is non-zero in more than one state of $V^0$, breaking the correspondence between monomials of $S^0$ and states of $V^0$. However, if the SPN is complete then $V^0 = V^1 = V^2$, and their states are in one-to-one correspondence. Therefore by the induction hypothesis the monomials of $V^1$ and $V^2$ are also in one-to-one correspondence and $S^0 = \sum_{x^0} (w_{01}\Phi^1(x^0) + w_{02}\Phi^2(x^0))\Pi(x^0)$; i.e., the expansion of $S^0$ is its network polynomial.

If $n^0$ is a product node, then $S^0 = \left(\sum_{x^1} \Phi^1(x^1)\Pi(x^1)\right)\left(\sum_{x^2} \Phi^2(x^2)\Pi(x^2)\right)$. If $V_1 \cap V_2 = \emptyset$, it follows immediately that the expansion of $V_0$ is its network polynomial. In the more general case, let $V_{12} = V_1 \cap V_2$, $V_{1-} = V_1 \setminus V_2$ and $V_{2-} = V_2 \setminus V_1$ and let $x^{12}$, $x^{1-}$ and $x^{2-}$ be the corresponding states. Since each $\Phi^1(x^1)$ is non-zero in exactly one state $x^1$ and similarly for $\Phi^2(x^2)$, each monomial in the product of $S^1$ and $S^2$ is nonzero in at most one state of $V^0 = V^1 \cup V^2$. If the SPN is not consistent, then at least one monomial in the product contains both the positive and negative indicators of a variable, $x_i$ and $\bar{x}_i$. Since no monomial in the network polynomial contains both $x_i$ and $\bar{x}_i$, this means the expansion of $S^0$ is not equal to its network polynomial. To ensure that each monomial in $S^0$ is non-zero in at least one state of $V_0$, for every $\Pi(x^{1-}, x^{12})$, $\Pi(x^{12}, x^{2-})$ pair there must exist a state $x^0 = (x^{1-}, x^{12}, x^{2-})$ where both $\Pi(x^{1-}, x^{12})$ and $\Pi(x^{12}, x^{2-})$ are 1, and therefore the indicators over $x^{12}$ in both monomials must be consistent. Since by the induction hypothesis they completely specify $x^{12}$, they must be the same in the two monomials. Therefore all $\Pi(x^{1-}, x^{12})$ and $\Pi(x^{12}, x^{2-})$ monomials must have the same $x^{12}$ indicators, i.e., the SPN must be consistent. □

Completeness and consistency are not necessary for validity; for example, the network $S(x_1, x_2, \bar{x}_1, \bar{x}_2) = \frac{1}{2}x_1x_2\bar{x}_2 + \frac{1}{2}x_1$ is incomplete and inconsistent, but satisfies $\Phi_S(e) = \sum_{x \in e} S(x)$ for all evidence $e$. However, completeness and consistency are necessary for the stronger property that every subnetwork of $S$ be valid. This can be proved by refutation. The input nodes are valid by definition. Let S be a node that violates either completeness or consistency but all of its descendants satisfy both conditions. We can show that S is not valid since it either undercounts the summation (if it is incomplete) or overcounts it (if it is inconsistent).

If an SPN $S$ is complete but inconsistent, its expansion includes monomials that are not present in its network polynomial, and $S(e) \geq \Phi_S(e)$. If $S$ is consistent but in-complete, some of its monomials are missing indicators relative to the monomials in its network polynomial, and $S(e) \leq \Phi_S(e)$. Thus invalid SPNs may be useful for approximate inference. Exploring this is a direction for future work.

Figure 2: A sum node $i$ can be viewed as the result of summing out a hidden variable $Y_i$; $y_{ij}$ represents the indicator $[Y_i = j]$ and $j$ ranges over the children of $i$.

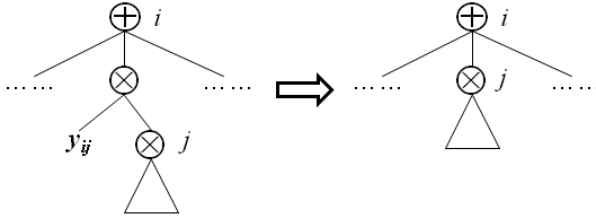

Completeness and consistency allow us to design deep architectures where inference is guaranteed to be efficient. This in turn makes learning them much easier.

**Definition 5** *An unnormalized probability distribution $\Phi(x)$ is* representable *by a sum-product network $S$ iff $\Phi(x) = S(x)$ for all states $x$ and $S$ is valid.*

$S$ then correctly computes all marginals of $\Phi(x)$, including its partition function.

**Theorem 2** *The partition function of a Markov network $\Phi(x)$, where $x$ is a $d$-dimensional vector, can be computed in time polynomial in $d$ if $\Phi(x)$ is representable by a sum-product network with a number of edges polynomial in $d$.*

*Proof.* Follows immediately from the definitions of SPN and representability. □

**Definition 6** *A sum-product network is* decomposable *iff no variable appears in more than one child of a product node.*

Decomposability is more restricted than consistency (e.g., $S(x_1, \bar{x}_1) = x_1 x_1$ is consistent but not decomposable.) This makes SPNs more general than representations that require decomposability, like arithmetic circuits [7], probabilistic context-free grammars [6], mixture models [32], junction trees [5], and others. (See also Section 3.)

SPNs can be extended to multi-valued discrete variables simply by replacing the Boolean indicators $[X_i = 1]$, $[X_i = 0]$ with indicators for the variable's possible values $x_i^j$: $[X_i = x_i^1], \ldots, [X_i = x_i^m]$, or $x_i^1, \ldots, x_i^m$ for short. For example, the multinomial distribution over $X_i$ is represented by $\sum_{j=1}^m p_i^j x_i^j$, where $p_i^j = P(X_i = x_i^j)$.

If an SPN $S$ is complete and consistent, and for every sum node $i$, $\sum_{j \in Ch(i)} w_{ij} = 1$, where $Ch(i)$ are the children of $i$, then $Z_S = 1$. In this case, we can view each sum node $i$ as the result of summing out an implicit hidden variable $Y_i$ whose values correspond to its children $Ch(i)$ (see Figure 2). This is because a variable is summed out by setting all its indicators to 1, and children of product nodes whose value is 1 can be omitted. Thus the SPN rooted at node $i$ can be viewed as a mixture model, with its children being the mixture components, which in turn are products of mixture models. If $i$ has no parent (i.e., it is the root), its children's weights are $Y_i$'s prior distribution: $w_{ij} = P(Y_i = j)$. Otherwise $w_{ij} = P(Y_i = j | \pi_i)$, where $\pi_i$ is the condition that, on at least one path from $Y_i$ to the root, all of $Y_i$'s ancestors have the values that lead to $Y_i$ (the ancestors being the hidden variables corresponding to the sum nodes on the path). If the network is also decomposable, the subnetwork rooted at the $j$th child then represents the distribution of the variables in it conditioned on $Y_i = j$. Thus an SPN can be viewed as a compact way to specify a mixture model with exponentially many mixture components, where subcomponents are composed and reused in larger ones. From this perspective, we can naturally derive an EM algorithm for SPN learning. (See Section 4.)

SPNs can be generalized to continuous variables by viewing these as multinomial variables with an infinite number of values. The multinomial's weighted sum of indicators $\sum_{j=1}^m p_i^j x_i^j$ then becomes the integral $\int p(x) dx$, where $p(x)$ is the p.d.f. of $X$. For example, $p(x)$ can be a univariate Gaussian. Thus SPNs over continuous variables have integral nodes instead of sum nodes with indicator children (or instead of indicators, since these can be viewed as degenerate sum nodes where one weight is 1 and the others are 0). We can then form sums and products of these nodes, as before, leading to a rich yet compact language for specifying high-dimensional continuous distributions. During inference, if the evidence includes $X = x$, the value of an integral node $n$ over $x$ is $p_n(x)$; otherwise its value is 1. Computing the probability of evidence then proceeds as before.

Given a valid SPN, the marginals of all variables (including the implicit hidden variables $Y$) can be computed by differentiation [7]. Let $n_i$ be an arbitrary node in SPN $S$, $S_i(x)$ be its value on input instance $x$, and $Pa_i$ be its parents. If $n_i$ is a product node, its parents (by assumption) are sum nodes, and $\partial S(x)/\partial S_i(x) = \sum_{k \in Pa_i} w_{ki} \partial S(x)/\partial S_k(x)$. If $n_i$ is a sum node, its parents (by assumption) are product nodes, and $\partial S(x)/\partial S_i(x) = \sum_{k \in Pa_i} (\partial S(x)/\partial S_k(x)) \prod_{l \in Ch_{-i}(k)} S_l(x)$, where $Ch_{-i}(k)$ are the children of the $k$th parent of $n_i$ excluding $n_i$. Thus we can evaluate $S_i$'s in an upward pass from input to the root, with parents following their children, and then compute $\partial S(x)/\partial w_{ij}$ and $\partial S(x)/\partial S_i(x)$ in a downward pass from the root to input, with children following parents. The marginals for the nodes can be derived from these partial derivatives [7]. In particular, if $n_i$ is a child of a sum node $n_k$, then $P(Y_k = i | e) \propto w_{ki} \partial S(e)/\partial S_k(e)$; if $n_i$ is an indicator $[X_s = t]$, then $P(X_s = t | e) \propto \partial S(e)/\partial S_i(e)$.

The continuous case is similar except that we have marginal densities rather than marginal probabilities.

The MPE state $\arg\max_{X,Y} P(X,Y|e)$ can be computed by replacing sums by maximizations. In the upward pass, a max node outputs the maximum weighted value among its children instead of their weighted sum. The downward pass then starts from the root and recursively selects the (or a) highest-valued child of a max node, and all children of a product node. Based on the results in Darwiche [7], we can prove that this will find the MPE state if the SPN is decomposable. Extension of the proof to consistent SPNs is straightforward since by definition no conflicting input indicators will be chosen. The continuous case is similar, and straightforward as long as computing the max and argmax of $p(x)$ is easy (as is the case with Gaussians).

## 3  SUM-PRODUCT NETWORKS AND OTHER MODELS

Let $R_M(D)$ be the most compact representation of distribution $D$ under moder class $M$, $\text{size}(R)$ be the size of representation $R$, $c > 0$ be a constant, and $\exp(x)$ be an exponential function. We say that model class $M_1$ is more general than model class $M_2$ iff for all distributions $D$ $\text{size}(R_{M_1}(D)) \leq c \cdot \text{size}(R_{M_2}(D))$ and there exist distributions for which $\text{size}(R_{M_2}(D)) \geq \exp(\text{size}(R_{M_1}(D)))$. In this sense, sum-product networks are more general than both hierarchical mixture models [32] and thin junction trees [5]. Clearly, both of these can be represented as SPNs without loss of compactness (see Figure 1). SPNs can be exponentially more compact than hierarchical mixture models because they allow mixtures over subsets of variables and their reuse. SPNs can be exponentially more compact than junction trees when context-specific independence and determinism are present, since they exploit these and junction trees do not. This holds even when junction trees are formed from Bayesian networks with context-specific independence in the form of decision trees at the nodes, because decision trees suffer from the replication problem [21] and can be exponentially larger than a DAG representation of the same function.

Figure 3 shows an SPN that implements a uniform distribution over states of five variables with an even number of 1's, as well as the corresponding mixture model. The distribution can also be non-uniform if the weights are not uniform. In general, SPNs can represent such distributions in size linear in the number of variables, by reusing intermediate components. In contrast, a mixture model (hierarchical or not) requires an exponential number of components, since each component must correspond to a complete state, or else it will assign non-zero probability to some state with an odd number of 1's.

Graphical models with junction tree clique potentials that cannot be simplified to polynomial size cannot be represented compactly as SPNs. More interestingly, as the previous example shows, SPNs can compactly represent some classes of distributions in which no conditional independences hold. Multi-linear representations (MLRs) also have this property [24]. Since MLRs are essentially expanded SPNs, an SPN can be exponentially more compact than the corresponding MLR.

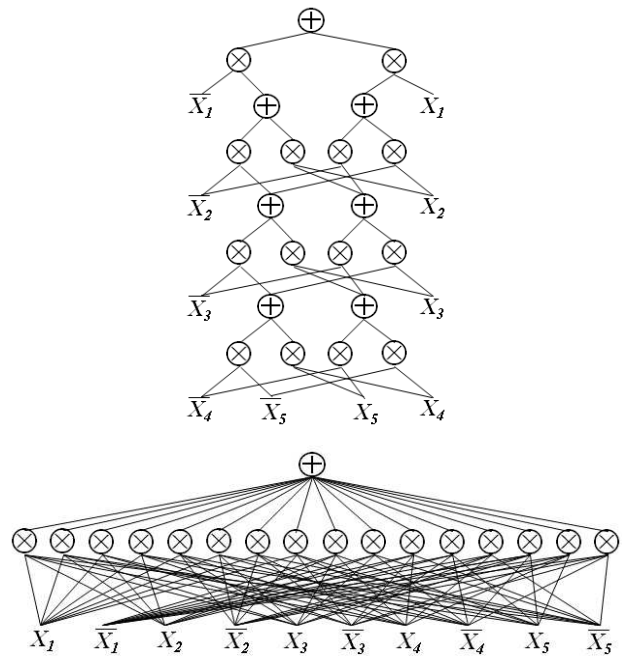

Figure 3: Top: SPN representing the uniform distribution over states of five variables containing an even number of 1's. Bottom: mixture model for the same distribution. For simplicity, we omit the uniform weights.

SPNs are closely related to data structures for efficient inference like arithmetic circuits [7] and AND/OR graphs [8]. However, to date these have been viewed purely as compilation targets for Bayesian network inference and related tasks, and have no semantics as models in their own right. As a result, the problem of learning them has not generally been considered. The two exceptions we are aware of are Lowd and Domingos [18] and Gogate et al. [12]. Lowd and Domingos's algorithm is a standard Bayesian network structure learner with the complexity of the resulting circuit as the regularizer, and does not have the flexibility of SPN learning. Gogate et al.'s algorithm learns Markov networks representable by compact circuits, but does not reuse subcircuits. Case-factor diagrams [19] are another compact representation, similar to decomposable SPNs. No algorithms for learning them or for computing the probability of evidence in them have been proposed to date.

We can view the product nodes in an SPN as forming a feature hierarchy, with the sum nodes representing distributions over them; in contrast, standard deep architectures explicitly represent only the features, and require the sums to be inefficiently computed by Gibbs sampling or otherwise approximated. Convolutional networks [15] alternate feature layers with pooling layers, where the pooling operation is typically max or average, and the features in each layer are over a subset of the input variables. Convolutional networks are not probabilistic, and are usually viewed as a vision-specific architecture. SPNs can be viewed as probabilistic, general-purpose convolutional networks, with average-pooling corresponding to marginal inference and max-pooling corresponding to MPE inference. Lee at al. [16] have proposed a probabilistic version of max-pooling, but in their architecture there is no correspondence between pooling and the sum or max operations in probabilistic inference, as a result of which inference is generally intractable. SPNs can also be viewed as a probabilistic version of competitive learning [27] and sigma-pi networks [25]. Like deep belief networks, SPNs can be used for nonlinear dimensionality reduction [14], and allow objects to be reconstructed from the reduced representation (in the case of SPNs, a choice of mixture component at each sum node).

**Algorithm 1** LearnSPN
   **Input:** Set $D$ of instances over variables $X$.
   **Output:** An SPN with learned structure and parameters.
   $S \leftarrow$ GenerateDenseSPN($X$)
   InitializeWeights($S$)
   **repeat**
     **for all** $d \in D$ **do**
        UpdateWeights($S$, Inference($S, d$))
     **end for**
   **until** convergence
   $S \leftarrow$ PruneZeroWeights($S$)
   **return** $S$

Probabilistic context-free grammars and statistical parsing [6] can be straightforwardly implemented as decomposable SPNs, with non-terminal nodes corresponding to sums (or maxes) and productions corresponding to products (logical conjunctions for standard PCFGs, and general products for head-driven PCFGs). Learning an SPN then amounts to directly learning a chart parser of bounded size. However, SPNs are more general, and can represent unrestricted probabilistic grammars with bounded recursion. SPNs are also well suited to implementing and learning grammatical vision models (e.g., [10, 33]).

## 4 LEARNING SUM-PRODUCT NETWORKS

The structure and parameters of an SPN can be learned together by starting with a densely connected architecture and learning the weights, as in multilayer perceptrons. Algorithm 1 shows a general learning scheme with online learning; batch learning is similar.

First, the SPN is initialized with a generic architecture. The only requirement on this architecture is that it be valid (complete and consistent). Then each example is processed in turn by running inference on it and updating the weights. This is repeated until convergence. The final SPN is obtained by pruning edges with zero weight and recursively removing non-root parentless nodes. Note that a weighted edge must emanate from a sum node and pruning such edges will not violate the validity of the SPN. Therefore, the learned SPN is guaranteed to be valid.

Completeness and consistency are general conditions that leave room for a very flexible choice of architectures. Here, we propose a general scheme for producing the initial architecture: 1. Select a set of subsets of the variables. 2. For each subset $R$, create $k$ sum nodes $S_1^R, \ldots, S_k^R$, and select a set of ways to decompose $R$ into other selected subsets $R_1, \ldots, R_l$. 3. For each of these decompositions, and for all $1 \leq i_1, \ldots, i_l \leq k$, create a product node with parents $S_j^R$ and children $S_{i_1}^{R_1}, \ldots, S_{i_l}^{R_l}$. We require that only a polynomial number of subsets is selected and for each subset only a polynomial number of decompositions is chosen. This ensures that the initial SPN is of polynomial size and guarantees efficient inference during learning and for the final SPN. For domains with inherent local structure, there are usually intuitive choices for subsets and decompositions; we give an example in Section 5 for image data. Alternatively, subsets and decompositions can be selected randomly, as in random forests [4]. Domain knowledge (e.g., affine invariances or symmetries) can also be incorporated into the architecture, although we do not pursue this in this paper.

Weight updating in Algorithm 1 can be done by gradient descent or EM. We consider each of these in turn.

SPNs lend themselves naturally to efficient computation of the likelihood gradient by backpropagation [26]. Let $n_j$ be a child of sum node $n_i$. Then $\partial S(x)/\partial w_{ij} = (\partial S(x)/\partial S_i(x))S_j(x)$ and can be computed along with $\partial S(x)/\partial S_i(x)$ using the marginal inference algorithm described in Section 2. The weights can then be updated by a gradient step. (Also, if batch learning is used instead, quasi-Newton and conjugate gradient methods can be applied without the difficulties introduced by approximate inference.) We ensure that $S(*) = 1$ throughout by renormalizing the weights at each step, i.e., projecting the gradient onto the $S(*) = 1$ constraint surface. Alternatively, we can let $Z = S(*)$ vary and optimize $S(X)/S(*)$.

SPNs can also be learned using EM [20] by viewing each sum node $i$ as the result of summing out a corresponding hidden variable $Y_i$, as described in Section 2. Now the inference in Algorithm 1 is the E step, computing the marginals of the $Y_i$'s, and the weight update is the M step, adding each $Y_i$'s marginal to its sum from the previous iterations and renormalizing to obtain the new weights.

In either case, MAP learning can be done by placing a prior on the weights. In particular, we can use a sparse prior, leading to a smaller SPN after pruning zero weights and thus to faster inference, as well as combatting overfitting.

Unfortunately, both gradient descent and EM as described above give poor results when learning deep SPNs. Gradient descent falls prey to the gradient diffusion problem: as more layers are added, the gradient signal rapidly dwindles to zero. This is the key difficulty in deep learning. EM also suffers from this problem, because its updates also become smaller and smaller as we go deeper. We propose to overcome this problem by using hard EM, i.e., replacing marginal inference with MPE inference. Algorithm 1 now maintains a count for each sum child, and the M step simply increments the count of the winning child; the weights are obtained by normalizing the counts. This avoids the gradient diffusion problem because all updates, from the root to the inputs, are of unit size. In our experiments, this made it possible to learn accurate deep SPNs, with tens of layers instead of the few typically used in deep learning.

## 5 EXPERIMENTS

We evaluated SPNs by applying them to the problem of completing images. This is a good test for a deep architecture, because it is an extremely difficult task, where detecting deep structure is key. Image completion has been studied quite extensively in graphics and vision communities (e.g., [31, 3]), but the focus tends to be restoring small occlusions (e.g., eyeglasses) to facilitate recognition tasks. Some recent machine learning works also showed selected image completion results [16, 1, 30], but they were limited and often focused on small images. In contrast, we conducted extensive evaluations where the half of each image is occluded.

We conducted our main evaluation on Caltech-101 [9], a well-known dataset containing images in 101 categories such as faces, helicopters, and dolphins. For each category, we set aside the last third (up to 50 images) for test and trained an SPN using the rest. For each test image, we covered half of the image and applied the learned SPN to complete the occlusion. Additionally, we also ran experiments on the Olivetti face dataset [28] containing 400 faces.

To initialize the SPN, we used an architecture that leverages local structure in image data. Specifically, in GenerateDenseSPN, all rectangular regions are selected, with the smallest regions corresponding to pixels. For each rectangular region, we consider all possible ways to decompose it into two rectangular subregions.

SPNs can also adopt multiple resolution levels. For example, for large regions we may only consider coarse region decompositions. In preliminary experiments, we found that this made learning much faster with little degradation in accuracy. In particular, we adopted an architecture that uses decompositions at a coarse resolution of $m$-by-$m$ for large regions, and finer decompositions only inside each $m$-by-$m$ block. We set $m$ to 4 in our experiments.

The SPNs learned in our experiments were very deep, containing 36 layers. In general, in our architecture there are $2(d-1)$ layers between the root and input for $d \times d$ images. The numbers for SPNs with multiple resolution levels can be computed similarly.

We used mini-batches in online hard EM; processing of instances in a batch can be trivially parallelized. Running soft EM after hard EM yielded no improvement. The best results were obtained using sums on the upward pass and maxes on the downward pass (i.e., the MPE value of each hidden variable is computed conditioning on the MPE values of the hidden variables above it and summing out the ones below). We initialized all weights to zero and used add-one smoothing when evaluating nodes. We penalized non-zero weights with an $L_0$ prior with parameter 1.[1]

To handle gray-scale intensities, we normalized the intensities of input images to have zero mean and unit variance, and treated each pixel variable $X_i$ as a continuous sample from a Gaussian mixture with $k$ unit-variance components. For each pixel, the intensities of training examples are divided into $k$ equal quantiles and the mean of each component is set to that of the corresponding quantile. We used four components in our experiments. (We also tried using more components and learning the mixing parameters, but it yielded no improvement in performance.)

We compared SPNs with deep belief networks (DBNs) [14] and deep Boltzmann machines (DBMs) [29]. These are state-of-the-art deep architectures and their codes are publicly available. DBNs and DBMs both consist of several layers of restricted Boltzmann machines (RBMs), but they differ in the probabilistic model and training procedure.

We also compared SPNs with principal component analysis (PCA) and nearest neighbor. PCA has been used extensively in previous image completion works [31]. We used 100 principal components in our experiments. (Results with higher or lower numbers are similar.) Despite its simplicity, nearest neighbor can give quite good results if an image similar to the test one has been seen in the past [13]. For each test image, we found the training image with most similar right (top) half using Euclidean distance, and returned its left (bottom) half.

We report mean square errors of the completed pixels of test images for these five algorithms. Table 1 show the average result among all Caltech-101 categories, as well as the results for a few example categories and Olivetti.[2] Note that the DBN results are not directly comparable with others. Using the original images and without additional preprocessing, the learned DBN gave very poor results, despite our extensive effort to experiment using the code from Hinton and Salakhutdinov [14]. Hinton and Salakhutdinov [14] reported results for image reconstruction on Olivetti faces, but they used reduced-scale images (25×25 compared to the original size of 64×64) and required a training set containing over 120,000 images derived via transformations like rotation, scaling, etc. By converting to the reduced scale and initializing with their learned model, the results improve significantly and so we report these results instead. Note that reducing the scale artificially lowers the mean square errors by reducing the overall variance. So although DBN appears to have lower errors than DBM and nearest neighbor, their completions are actually much worse (see examples in Figure 5).

---

[1] Hard EM permits the use of an $L_0$ prior, which can be incorporated in finding the MAP state. For gradient descent, an $L_1$ prior was used instead.

Table 1: Mean squared errors on completed image pixels in the left or bottom half. NN is nearest neighbor.

| LEFT | SPN | DBM | DBN | PCA | NN |
|---|---|---|---|---|---|
| Caltech (ALL) | 3551 | 9043 | 4778 | 4234 | 4887 |
| Face | 1992 | 2998 | 4960 | 2851 | 2327 |
| Helicopter | 3284 | 5935 | 3353 | 4056 | 4265 |
| Dolphin | 2983 | 6898 | 4757 | 4232 | 4227 |
| Olivetti | 942 | 1866 | 2386 | 1076 | 1527 |

| BOTTOM | SPN | DBM | DBN | PCA | NN |
|---|---|---|---|---|---|
| Caltech (ALL) | 3270 | 9792 | 4492 | 4465 | 5505 |
| Face | 1828 | 2656 | 3447 | 1944 | 2575 |
| Helicopter | 2801 | 7325 | 4389 | 4432 | 7156 |
| Dolphin | 2300 | 7433 | 4514 | 4707 | 4673 |
| Olivetti | 918 | 2401 | 1931 | 1265 | 1793 |

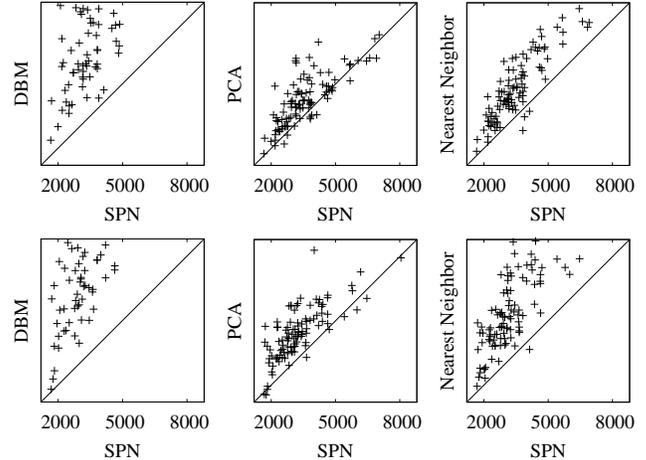

Figure 4: Scatter plots comparing SPNs with DBMs, PCA, and nearest neighbor in mean square errors on Caltech-101. Each point represents an image category. The $y$ axes have the same scale as the $x$ axes. Top: left completion. Bottom: bottom completion.

Overall, SPN outperforms all other methods by a wide margin. PCA performs surprisingly well in terms of mean square errors compared to methods other than SPN, but their completions are often quite blurred since they are a linear combination of prototypical images. Nearest neighbor can give good completions if there is a similar image in training, but in general their completions can be quite poor. Figure 4 shows the scatter plots comparing SPNs with DBMs, PCA, and nearest neighbor, which confirms the advantage of SPN. The differences are statistically significant by the binomial sign test at the $p < 0.01$ level.

Compared to state-of-the-art deep architectures [14, 16, 29], we found that SPNs have three significant advantages.

First, SPNs are considerably simpler, theoretically more well-founded, and potentially more powerful. SPNs ad-mit efficient exact inference, while DBNs and DBMs require approximate inference. The problem of gradient diffusion limits most learned DBNs and DBMs to a few layers, whereas with online hard EM, very deep accurate SPNs were learned in our experiments. In practice, DBNs and DBMs also tend to require substantially more engineering. For example, we set the hyperparameters for SPNs in preliminary experiments and found that these values worked well for all datasets. We also used the same architecture throughout and let learning adapt the SPN to the details of each dataset. In contrast, DBNs and DBMs typically require a careful choice of parameters and architectures for each dataset. (For example, the default learning rate of 0.1 leads to massive divergence in learning Caltech images with DBNs.) SPN learning terminates when the average log-likelihood does not improve beyond a threshold (we used 0.1, which typically converges in around 10 iterations; 0.01 yielded no improvement in initial experiments). For DBNs/DBMs, however, the number of iterations has to be determined empirically using a large development set. Further, successful DBN/DBM training often requires extensive preprocessing of the examples, while we used essentially none for SPNs.

---

[2]The complete set of results and the SPN code will be available for download at http://alchemy.cs.washington.edu/spn.

Second, SPNs are at least an order of magnitude faster in both learning and inference. For example, learning Caltech faces takes about 6 minutes with 20 CPUs, or about 2 hours with one CPU. In contrast, depending on the number of learning iterations and whether a much larger transformed dataset is used (as in [14, 29]), learning time for DBNs/DBMs ranges from 30 hours to over a week. For inference, SPNs took less than a second to find the MPE completion of an image, to compute the likelihood of such a completion, or to compute the marginal probability of a variable, and all these results are exact. In contrast, esti-potential of SPNs for object recognition. Lee et al. [16] reported results for convolutional DBNs (CDBNs) by training a CDBN for each of the three Caltech-101 categories faces, motorbikes, and cars, and then computed area under precision-recall curve (AUC) by comparing the probabilities for positive and negative examples in each classification problem (one class vs. others). We followed their experimental setting and conducted experiments using SPNs. Table 2 compares the results with those obtained using top layer features in convolutional DBNs (CDBNs) (see Figure 4 in [16]). SPNs obtained almost perfect results in all three categories whereas CDBNs' results are substantially lower, particularly in motorbikes and cars.[4]

Figure 5: Sample face completions. Top to bottom: original, SPN, DBM, DBN, PCA, nearest neighbor. The first three images are from Caltech-101, the rest from Olivetti.

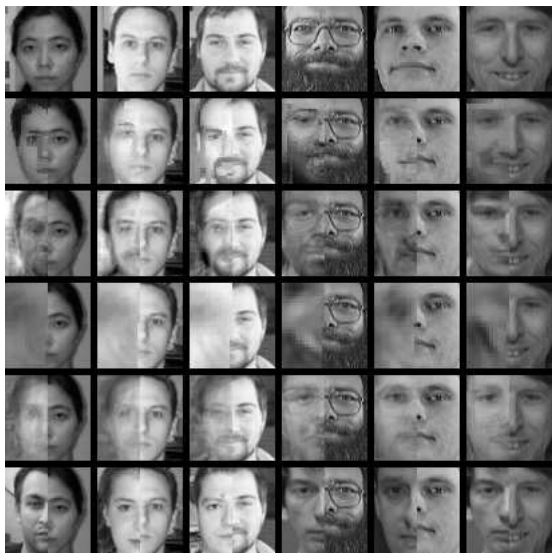

Table 2: Comparison of the area under the precision-recall curve for three classification problems (one class vs. the other two).

| Architecture | Faces | Motorbikes | Cars |
| --- | --- | --- | --- |
| SPN | 0.99 | 0.99 | 0.98 |
| CDBN (top layer) | 0.95 | 0.81 | 0.87 |

mating likelihood in DBNs or DBMs is a very challenging problem [30]; estimating marginals requires many Gibbs sampling steps that may take minutes or even hours, and the results are approximate without guarantee on the quality.

Third, SPNs appear to learn much more effectively. For example, Lee et al. [16] show five faces with completion results in Figure 6 of their paper. Their network was only able to complete a small portion of the faces, leaving the rest blank (starting with images where the visible side already contained more than half the face).[3] The completions generated by DBMs look plausible in isolation, but they are often at odds with the observed portion and the same completions are often reused in different images. The mean square error results in Table 1 confirmed that the DBM completions are often not very good. Among all categories, DBMs performed relatively well in Caltech and Olivetti faces. So we contrast example completions in Figure 5, which shows the results for completing the left halves of previously unseen faces. The DBM completions often seem to derive from the nearest neighbor according to its learned model, which suggests that they might not have learned very deep regularities. In comparison, the SPN successfully completed most faces by hypothesizing the correct locations and types of various parts like hair, eye, mouth, and face shape and color. On the other hand, the SPN also has some weaknesses. For example, the completions often look blocky.

We also conducted preliminary experiments to evaluate the

## 6 SUM-PRODUCT NETWORKS AND THE CORTEX

The cortex is composed of two main types of cells: pyramidal neurons and stellate neurons. Pyramidal neurons excite the neurons they connect to; most stellate neurons inhibit them. There is an interesting analogy between these two types of neurons and the nodes in SPNs, particularly when MAP inference is used. In this case the network is composed of max nodes and sum nodes (logs of products). (Cf. Riesenhuber and Poggio [23], which also uses max and sum nodes, but is not a probabilistic model.) Max nodes are analogous to inhibitory neurons in that they select the highest input for further propagation. Sum nodes are analogous to excitatory neurons in that they compute a sum of their inputs. In SPNs the weights are at the inputs of max nodes, while the analogy with the cortex suggests having them at the inputs of sum (log product) nodes. One can be mapped to the other if we let max nodes ignore their children's weights and consider only their values. Possible justifications for this include: (a) it potentially reduces computational cost by allowing max nodes to be merged; (b) ignoring priors may improve discriminative performance [11]; (c) priors may be approximately encoded by the number of units representing the same pattern, and this may facilitate online hard EM learning. Unlike SPNs, the cortex has no single root node, but it is straighforward to extend SPNs to have multiple roots, corresponding to simultaneously computing multiple distributions with shared structure. Of course, SPNs are still biologically unrealistic in many ways, but they may nevertheless provide an interesting addition to the computational neuroscience toolkit.

---

[3]We were unable to obtain their code for head-to-head comparison. We should note that the main purpose of their figure is to illustrate the importance of top-down inference.

[4]We should note that the main point of their results is to show that features in higher layers are more class-specific.

# 7 CONCLUSION

Sum-product networks (SPNs) are DAGs of sums and products that efficiently compute partition functions and marginals of high-dimensional distributions, and can be learned by backpropagation and EM. SPNs can be viewed as a deep combination of mixture models and feature hierarchies. Inference in SPNs is faster and more accurate than in previous deep architectures. This in turn makes learning faster and more accurate. Our experiments indicate that, because of their robustness, SPNs require much less manual engineering than other deep architectures. Much remains to be explored, including other learning methods for SPNs, design principles for SPN architectures, extension to sequential domains, and further applications.

**Acknowledgements** We thank Ruslan Salakhutdinov for help in experiments with DBNs. This research was partly funded by ARO grant W911NF-08-1-0242, AFRL contract FA8750-09-C-0181, NSF grant IIS-0803481, and ONR grant N00014-08-1-0670. The views and conclusions contained in this document are those of the authors and should not be interpreted as necessarily representing the official policies, either expressed or implied, of ARO, AFRL, NSF, ONR, or the United States Government.